\documentclass[a4paper, 10pt, journal]{IEEEtran}

\usepackage{color}
\usepackage{comment}
\usepackage{colortbl}  
\usepackage{xcolor}
\usepackage{epsfig}
\usepackage{graphicx}
\usepackage{bbding}
\usepackage{array, caption, threeparttable}
\usepackage{amsmath}
\usepackage{amssymb}
\usepackage{comment}
\usepackage{multirow}

\usepackage{hyperref}
\hypersetup{hidelinks}

\usepackage{booktabs}
\usepackage{array, caption, threeparttable}
\usepackage{caption}
\usepackage{subfigure}
\usepackage[ruled]{algorithm2e}
\usepackage{listings}
\usepackage{color}
\usepackage{mathtools}

\usepackage{todonotes}
\usepackage{gensymb}
\usepackage{setspace}

\usepackage{algorithmic}

\title{\LARGE \bf
SFDA-rPPG: Source-Free Domain Adaptive Remote Physiological Measurement with Spatio-Temporal Consistency
}

\author{Yiping Xie, Zitong Yu, Bingjie Wu, Weicheng Xie and Linlin Shen
\thanks{This work was supported by Open Fund of National Engineering Laboratory for Big Data System Computing Technology (Grant No. SZU-BDSC-OF2024-02) and National Natural Science Foundation of China under Grant 62306061. Corresponding authors: Zitong Yu (email: zitong.yu@ieee.org) and Weicheng Xie (email: wcxie@szu.edu.cn).}
\thanks{Y. Xie, W. Xie, and L. Shen are with Computer Vision Institute, School of Computer Science \& Software Engineering, Shenzhen Institute of Artificial Intelligence and Robotics for Society, Guangdong Key Laboratory of Intelligent Information Processing, Shenzhen University, Shenzhen, 518060.}
\thanks{Z. Yu is with School of Computing and Information Technology, Great Bay University, Dongguan, 523000, and National Engineering Laboratory for Big Data System Computing Technology, Shenzhen University, Shenzhen 518060}
\thanks{B. Wu is with Institute of High Performance Computing (IHPC), Agency for Science, Technology and Research (A*STAR), Singapore.}}

\begin{document}

\maketitle

\begin{abstract}

Remote Photoplethysmography (rPPG) is a non-contact method that uses facial video to predict changes in blood volume, enabling physiological metrics measurement. Traditional rPPG models often struggle with poor generalization capacity in unseen domains. Current solutions to this problem is to improve its generalization in the target domain through Domain Generalization (DG) or Domain Adaptation (DA). However, both traditional methods require access to both source domain data and target domain data, which cannot be implemented in scenarios with limited access to source data, and another issue is the privacy of accessing source domain data. In this paper, we propose the first Source-free Domain Adaptation benchmark for rPPG measurement (SFDA-rPPG), which overcomes these limitations by enabling effective domain adaptation without access to source domain data. Our framework incorporates a Three-Branch Spatio-Temporal Consistency Network (TSTC-Net) to enhance feature consistency across domains. Furthermore, we propose a new rPPG distribution alignment loss based on the Frequency-domain Wasserstein Distance (FWD), which leverages optimal transport to align power spectrum distributions across domains effectively and further enforces the alignment of the three branches. Extensive cross-domain experiments and ablation studies demonstrate the effectiveness of our proposed method in source-free domain adaptation settings. Our findings highlight the significant contribution of the proposed FWD loss for distributional alignment, providing a valuable reference for future research and applications. The source code is available at \href{https://github.com/XieYiping66/SFDA-rPPG}{https://github.com/XieYiping66/SFDA-rPPG}. 


\end{abstract}

\section{INTRODUCTION}

Through the change of blood volume in the optical information of facial video, rPPG (Remote Photoplethysmography) technology can estimate physiological metrics such as heart rate, respiratory rate and so on. Early methods estimate the rPPG signal by subtle color changes~\cite{verkruysse2008remote, balakrishnan2013detecting, poh2010non, tulyakov2016self, de2013robust, wang2016algorithmic, park2018remote} in the face region, which are extracted from frames by using face detection methods such as MTCNN~\cite{zhang2016joint}. However, these methods need to manually set the region of interest and some filtering operations, and are easily influenced by factors that have an impact on skin color, such as illumination changes, and lack good robustness.

With the development of deep learning, rPPG-based on deep learning can overcome the influence of these non-physiological factors~\cite{nowara2021benefit, yu2020autohr,yu2021transrppg, yu2019remote1, vspetlik2018visual}.For example, use 2D-CNN ~\cite{vspetlik2018visual} and 3D-CNN ~\cite{yu2019remote1} to optimize training and learn richer rPPG information. Besides, Sun et al. propose an unsupervised contrastive learning framework using the spatiotemporal similarity of rPPG to train the model to learn richer rPPG knowledge without the access to labels. However, they may fail when dealing with unseen domains or encountering different domains with domain gaps, which are common in real application scenarios, such as domain gaps caused by different data sets collection methods. How to achieve better cross-domain performance is one of the current hot topics in rPPG research.

\begin{figure}
\centering
\includegraphics[width=\linewidth, height=7cm]{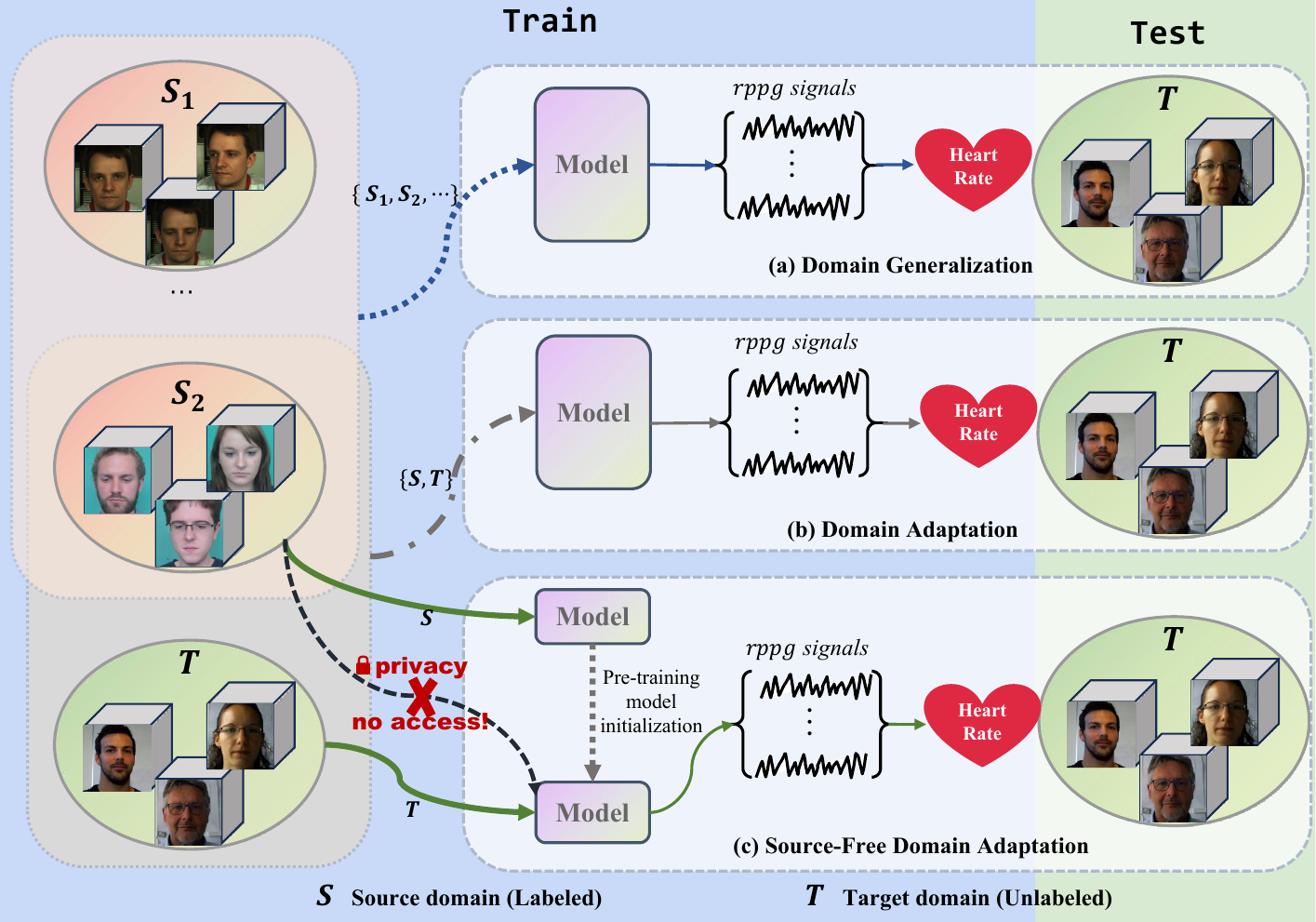}
\caption{Overview of cross-domain methods. (a) Multi-source domain generalization~\cite{lu2023neuron} utilizes labeled data from multiple source domains. (b) Unsupervised domain adaptation~\cite{du2023dual} leverages labeled data from both the source and target domains. (c) Source-free domain adaptation utilizes a pretrained model from the source domain and unlabeled data from the target domain. }
\label{fig:domain adaptation}
\end{figure}

Currently, the domain adaptation (DA) and the domain generalization (DG) technologies are helpful to improve the domain gap problem caused by the change of non-physiological factors~\cite{ganin2016domain, li2020maximum}. However, due to the unclear example-specific differences, this solution may not work well directly for rPPG measurement~\cite{wang2023hierarchical}. For example, many general domain adaptation methods use clustering to obtain more accurate category pseudo-labels, but in rPPG, the signal cannot be clustered unless it is categorized and broken down into different heart rate segments, which is obviously less accurate than the intuitive category. In the recent research of rPPG, Chung et al.~\cite{chung2022domain} improve rPPG estimation in unseen domain by using the feature learning framework of domain replacement and domain expansion. Du et al.~\cite{du2023dual} align the middle domain with a dual-bridge network and synthesize the target noise in the source domain to reduce the domain difference. However, these methods all need to use source domain data, leading to privacy problems. That is, both source domain and target domain data are used for training. But is there a way to adapt a model on a target domain based solely on previous knowledge, i.e., a model trained on the source domain, without accessing the source domain data during adaptation? This approach can effectively protect the privacy of source domain data, since only the model learned in the source domain needs to be used during adaptation. Therefore, we introduce Source-free domain adaptation paradigm in this paper.

As shown in Fig. \ref {fig:domain adaptation}, we introduce the essential differences between different cross-domain methods. Domain generalization-based rPPG methods~\cite{lu2023neuron} seeks to enhance model generalizability by leveraging information from multiple source domains, as depicted in Fig. \ref {fig:domain adaptation}(a). The DG method uses multiple source domain data for learning at the same time to improve the generalization of the model. However, it does not use the data of the target domain, but only learns the general knowledge in different fields. Traditional unsupervised DA-based rPPG aim to minimize the discrepancy between source and target domains by examining relationships between their respective datasets, as illustrated in Fig. \ref {fig:domain adaptation}(b). Ordinary DA methods use both source domain data and target domain data and cannot protect the privacy of either domain's data. These methods inherently utilize source domain data during training, raising privacy concerns for facial data in rPPG applications. Conversely, our approach employs source-free domain adaption (SFDA) based methods, as depicted in Fig. \ref {fig:domain adaptation}(c), which relies solely on a pre-trained model from source domain data and incorporates target domain’s unlabeled data for domain adaptation. At the same time, our method adapts the knowledge learned by the model trained on the source domain data to another unlabeled domain, which is the same as the idea of people gradually learning in the real world, such as using mathematical knowledge to solve physics. Our SFDA-rPPG framework integrates a Three-Branch Spatio-Temporal Consistency Network (TSTC-Net) to improve feature consistency across different domains. In addition, we propose a globally distributed alignment loss that is more appropriate for the rPPG task. Wasserstein distance loss is used in previous work~\cite{sun2022non} to leverage the timing of systolic peaks from contact PPG as labels for model training. However, in~\cite{sun2022non}, it is mainly applied in the time domain, and rPPG signals in the time domain are usually affected by various noises, which may be separated into high or low frequency regions in the frequency domain. Therefore, theoretically, WD loss in the frequency domain (FWD) focuses more on the power spectral density distribution across different frequency bands, and these distributions are more stable compared to rPPG signals in the time domain. So we use FWD loss for alignment in this stucy. In order to verify its effectiveness, we also compare the performence of Time-domain WD and Frequency-domain WD in Sec.\ref{sec:Discussion and Visualization}. 

Our main contributions are summarized as follows:

\begin{itemize}
    \item We propose a novel SFDA-rPPG framework, incorporating a Three-Branch Spatio-Temporal Consistency Network (TSTC-Net) to enhance rPPG measurement across domains. To the best of our knowledge, this is the first application of a source-free domain adaptation (SFDA) approach in the rPPG field, ensuring both robust performance and improved privacy and security.
    
    \item We propose a globally aligned Frequency-domain Wasserstein Distance (FWD) loss for rPPG distribution learning which captures more stable power spectral density distributions across frequency bands.

    \item We conduct comprehensive cross-dataset examinations to demonstrate the exceptional performance of SFDA-rPPG along with the efficacy of WD loss.
    
\end{itemize}

\section{Related Work}

\subsection{rPPG measurement}
As the rPPG signal change is very subtle, it is difficult to directly achieve from videos. Some traditional methods based on face and mathematical model analysis have been proposed. For example, Verkruysse et al.~\cite{balakrishnan2013detecting} use ambient light to recover rPPG from facial skin area for the first time. And then, PCA~\cite{balakrishnan2013detecting}, ICA~\cite{poh2010non} and other methods are used to decompose the original time signal to obtain rPPG signal, and the adaptive matrix~\cite{tulyakov2016self} is used to reduce noise and capture consistent clues related to rPPG. And color subspace transformation methods such as CHROM~\cite{de2013robust} based on chromaticity and POS~\cite{wang2016algorithmic} based on orthogonal projection plane of skin color are used to generate HR waveform. Later, Park, Sang Bae, et al. propose using skin boundary filter and mask filter to impove the rPPG performance~\cite{park2018remote}. 

However, the illumination conditions and motions are complicated, making manual traditional methods difficult to implement. Therefore, many deep learning methods have been proposed, including supervised and unsupervised methods. For supervised methods, ŠPetlík et al.~\cite{vspetlik2018visual} measure rPPG signals through 2D-CNN for the first time, and conduct end-to-end training through alternating optimization. And Yu et al. propose an end-to-end spatio-temporal network (PhysNet~\cite{yu2019remote1}), which considers the temporal context in facial video, and the first transformer framework (TranRPPG~\cite{yu2021transrppg}), which enhances automatic rPPG ferture representation on 3D mask face task. For unsupervised methods, the temporal and spatial characteristics of rPPG are usually used for learning. For example, the contrastive learning framework~\cite{sun2022contrast} proposed by Sun et al. explored the spatial and temporal similarities of rPPG. A 3DCNN model is used to generate multiple rPPG signals from each video at different spatiotemporal locations, and the rPPG signals from the same video are pulled together, while the rPPG signals from different videos are pushed apart. These methods can achieve good performance on a single dataset.

Nowadays, researchers are increasingly focusing on cross-dataset performance. Most current methods test the performance across different datasets. We also refer to data from different datasets as different domains because different rPPG data have different lighting conditions and other realistic influencing factors. In the recent research of rPPG, Chung et al.~\cite{chung2022domain} improve rPPG estimation in unseen domain by using the feature learning framework of domain replacement and domain expansion. Du et al.~\cite{du2023dual} align the middle domain with a dual-bridge network and synthesize the target noise in the source domain to reduce the domain difference. They all achieve good performance in cross-domain situations.

\begin{figure*}
\centering
\includegraphics[scale=0.5]{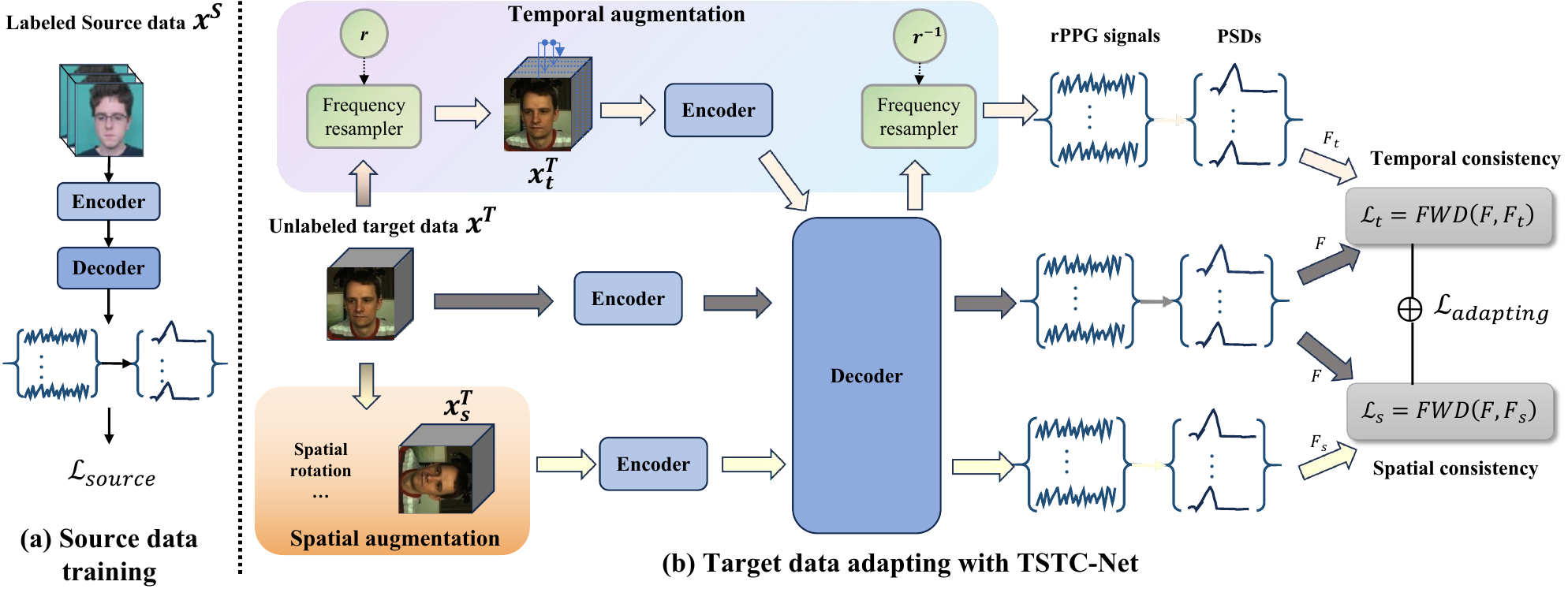}
\vspace{-0.9em}
  \caption{\small{
  The proposed SFDA-rPPG method consists of two distinct stages. Firstly, the pre-training stage involves pre-training the model with labeled source domain data. Secondly, the adaptive stage utilizes a spatio-temporal branching structure for consistency learning, enabling the source model to adapt more effectively to the unlabeled target domain data. The encoder and decoder are disentangled from PhysNet~\cite{yu2019remote1} to facilitate improved representation learning.}
  }
 
\label{fig:network}
\vspace{-1.1em}
\end{figure*}

\subsection{Source-Free Domain Adaptation}
In terms of Source-FreeDomainAdaptation (SFDA), research has made significant progress in recent years. The goal of SFDA is to adapt a model trained on the source domain by using only the target domain data without accessing the source domain data. This has important implications in terms of privacy protection and data transmission costs. The original SFDA methods mostly focused on adjusting model parameters through pseudo-label generation of target domain data. For example, Liang et al.~\cite{liang2020we} proposed a passive domain transfer method that uses pseudo labels for domain adaptation to improve adaptability by maximizing the entropy of target domain data. Another direction is to perform passive domain transfer through generative adversarial networks (GANs). Kundu et al.~\cite{kundu2020universal} proposed a GAN-based SFDA method to improve the domain adaptability of the model through adversarial generation and training. In the video field, Zhuang et al.~\cite{li2023source} proposed a Source-Free video domain adaptation method that utilizes space-time-history consistency learning. This method effectively utilizes the spatiotemporal consistency of the target domain video data and significantly improves Adaptive performance on video data. Furthermore, recent research has also explored implementing SFDA through feature alignment. For example, Yeh et al.~\cite{xia2021adaptive} proposed a source-free domain transfer method for feature alignment by maximizing the consistency of source and target domain features, which demonstrated excellent performance on multiple benchmark datasets. They all achieve effective adaptation to the target domain without the need for source domain data through different technical means. This provides new solutions for scenarios that face data privacy or data acquisition difficulties in practical applications.


\section{Methodology}

\subsection{Problem Definition}

Same as other DA and DG tasks, we employ $D_S=\left\{\left(X_{i}^{S}, Y_{i}^{S}\right)\right\}_{i=1}^{n_{S}}$ to define the source-domain dataset, where $n_S$ signifies the number of labeled source-domain data. Here, $X_i^S \in \mathbb{R}^{t \times w \times h \times 3}$ represents a video sample. The label of the source-domain denoted as $Y_i^S = [y_{1}, y_{2},…, y_{n}] $ where $y_{j} \in \mathbb{R}$, embodies the photoplethysmography (PPG) signal. Besides, we let $D_T = \{X_i^T\}_{i=1}^{n_T}$ signify the target-domain dataset, where $n_T$ represents the number of samples in the target domain, and the target-domain data is also characterized by $X_i^T \in \mathbb{R}^{t \times w \times h \times 3}$. It should be noted that our target domain data is unlabeled, which means we need to make the model well adapted to the unlabeled target data in the adaptation phase. At the same time, throughout the adaptation phase, our approach prohibits access to the source domain data $D_S$, relying solely on the unlabeled target domain data $D_T$ and the pre-trained source domain model. The primary objective is to mitigate the domain discrepancy through training, thereby enhancing the performance of the model encoder $E$ and decoder $D$ in the target domain.

\begin{algorithm}[t]
	\renewcommand{\algorithmicrequire}{\textbf{Input:}}
	\renewcommand{\algorithmicensure}{\textbf{Output:}}
	\caption{SFDA-rPPG adapting process}
	\label{alg1}
	\begin{algorithmic}[1]
     \REQUIRE unlabeled target data $\{X_i^T\}_{i=1}^{n_T}$, pretrained model weights $H_w=\mathcal{D}_w \circ E_w$, and total epoch number $e$
		\STATE  Initialize encoders $E$, $E_s$, $E_t$ with pre-training weights $E_w$, and initialize decoder $\mathcal{D}$ with $\mathcal{D}_w$;
            \FOR{$i=1$ to $e$}
            \STATE Get traget mini-batch $X_i^T$. 
            \\ \texttt{\# Spatial augmentation}.
            \STATE Get the spatial augmentation version $X_s^T$ with Eq.\eqref{eq:spatialequation}.
            \STATE Calculate spacial prediction signals $f_s = \mathcal{D}(E_s(x_s^T))$  and its PSD $F_s$.
            \\ \texttt{\# Temporal augmentation}.
            \STATE Generate random number $r$ $\in$ [0.66,0.80].
            \STATE Calculate the temporal prediction signals $f_t$ by frequency resampler with Eq.\eqref{eq:temporalequation}.
            \STATE Transform $f_t$ into frequency domain to get its PSD $F_t$.
            \\ \texttt{\# Obtain original data}.
            \STATE Calculate the original data prediction signals $f = \mathcal{D}(E(x_i^T)$ and its PSD $F$.
            \STATE Calculate FWD consistency loss $\mathcal{L}_{adapting}$ according to Eq.\eqref{eq:consistencyloss}.
            \STATE Update $E$,$E_s$,$E_t$,$\mathcal{D}$ with $\mathcal{L}_{adapting}$.
            \ENDFOR
	\end{algorithmic}  
\end{algorithm}

\subsection{Overview of SFDA-rPPG}

Inspired by the work~\cite{li2023source}, we introduce the spatio-temporal consistency network for SFDA-rPPG. As illustrated in Fig.~\ref{fig:network}, this network leverages the learned model weights $H_w$ from the source domain and the unlabeled target data $D_T$ to facilitate domain adaptation. Central to our method is the master encoder $E$ trained within the model, alongside the temporal and spatial branch encoders $E_t$ and $E_s$, and the decoder $D$. Overall, our approach unfolds in two main stages: 

(1) During the source-domain pre-training phase, the labeled source-domain data guides the model training to acquire the model weights $ H_w $, where $ H_w = \mathcal{D}_w \circ E_w $ with $ \mathcal{D}_w $ and $ E_w $ representing the pre-trained encoder and decoder weights of the model. 

(2) Subsequently, in the target domain adaptation phase, we adopt a Three-Branch Spatio-Temporal Consistency Network (TSTC-Net). Initially, the pre-trained model’s encoders and decoders serve as starting parameters, which are then fine-tuned on the target data to train the target encoders $ E $, $ E_s $, $ E_t $, and decoder $ \mathcal{D} $.

We also provide the detailed algorithm of the adaptation stage in Algorithm \ref{alg1}.

\subsection{Three-branch Spatio-Temporal Consistency Network}

To encourage the model to acquire stable predictive capabilities from the spatial and temporal transformations, in target domain adaptation, we partition the data into three distinct streams: (1) unlabeled target data, (2) spatial augmentation data, and (3) temporal augmentation data. We posit that an effective model should exhibit robustness across identical samples, yielding consistent outcomes for both the original sample and its temporally and spatially augmented variants. Consequently, we consistently align the outputs of the temporal and spatial branches with those of the unaltered target data branch. Also, to prevent the model from learning ambiguous feature representations, we intend for the mid- branch encoder $E$ to acquire a comprehensive representation and $E_s$,  $E_t$ to respectively acquire spatial and temporal representation. Subsequently, a shared decoder is utilized to learn task-specific feature decoding.

\textbf{Spatial augmentation.}\quad Following~\cite{yue2023facial}, we randomly apply one of six distinct spatial transformations to the video sample frames. Specifically, we establish the augmentation method set ${AU}_S$, including six spatial transformations: image rotations (0°, 90°, 180°, and 270°), and horizontal and vertical flips. We randomly choose an augmentation from the set and implement it on the current video sample frame sequence denoted as $X_i^T=\{x_1,x_2,x_3,…\}$, representing the $i$-th sequence of video frames in the target dataset,
\begin{equation}
\label{eq:spatialequation}
X_s^T=\{{AU}_{s,j} (x_1 ), {AU}_{s,j} (x_2 ), {AU}_{s,j} (x_3 ),… \},
\end{equation}
where ${AU}_{s,j}$ is the j-th spatial augmentation randomly selected in ${AU}_{s}$, and j belongs to the interval [1,6]. Then, we get the predicted rPPG signal through $ f_s = \mathcal{D}(E_s(X_s^T)) $ and power spectral density (PSD) $F_s$.

\textbf{Temporal augmentation.}\quad The method of directly masking off some frame sequences for time augmentation is not effective~\cite{shao2023tranphys} in the field of rPPG. Instead, we employ the frequency resampler introduced in~\cite{gideon2021way} to generate our temporal augmented samples. In a similar way, we randomly select the resampling factor $r$ within the range of [0.66, 0.80]. The difference is that we don't need to generate negative sample pairs, so we utilize the resampling mechanisms twice in succession directly. Initially, the sample is resampled by the $r$ resampling factor to generate a sample with higher heart rate. Then, to generate the prediction signal of the temporal augmented samples with the required scale, the signal value obtained after model estimation directly uses the inverse of $r$ as a new factor to re-sample the result back to the original scale. This process adheres to the following formula,
\begin{equation}
\label{eq:temporalequation}
f_t={FR}_{r^{-1} } (\mathcal{D}(E_t ({FR}_r (X_i^t )))),
\end{equation}
where ${FR}_r$ is a frequency resampler with $r$ as resampling factor and ${FR}_{r^{-1} }$ is a frequency resampler with the factor of the inverse of $r$. This specific method will directly obtain the rPPG signal $f_t$ of the temporal augmentation branch, and then its PSD $F_t$ can be obtained.

\subsection{Freqiemcy-domain Wasserstein Distance Loss}
\begin{figure}[t]
\centering
\includegraphics[scale=0.3]{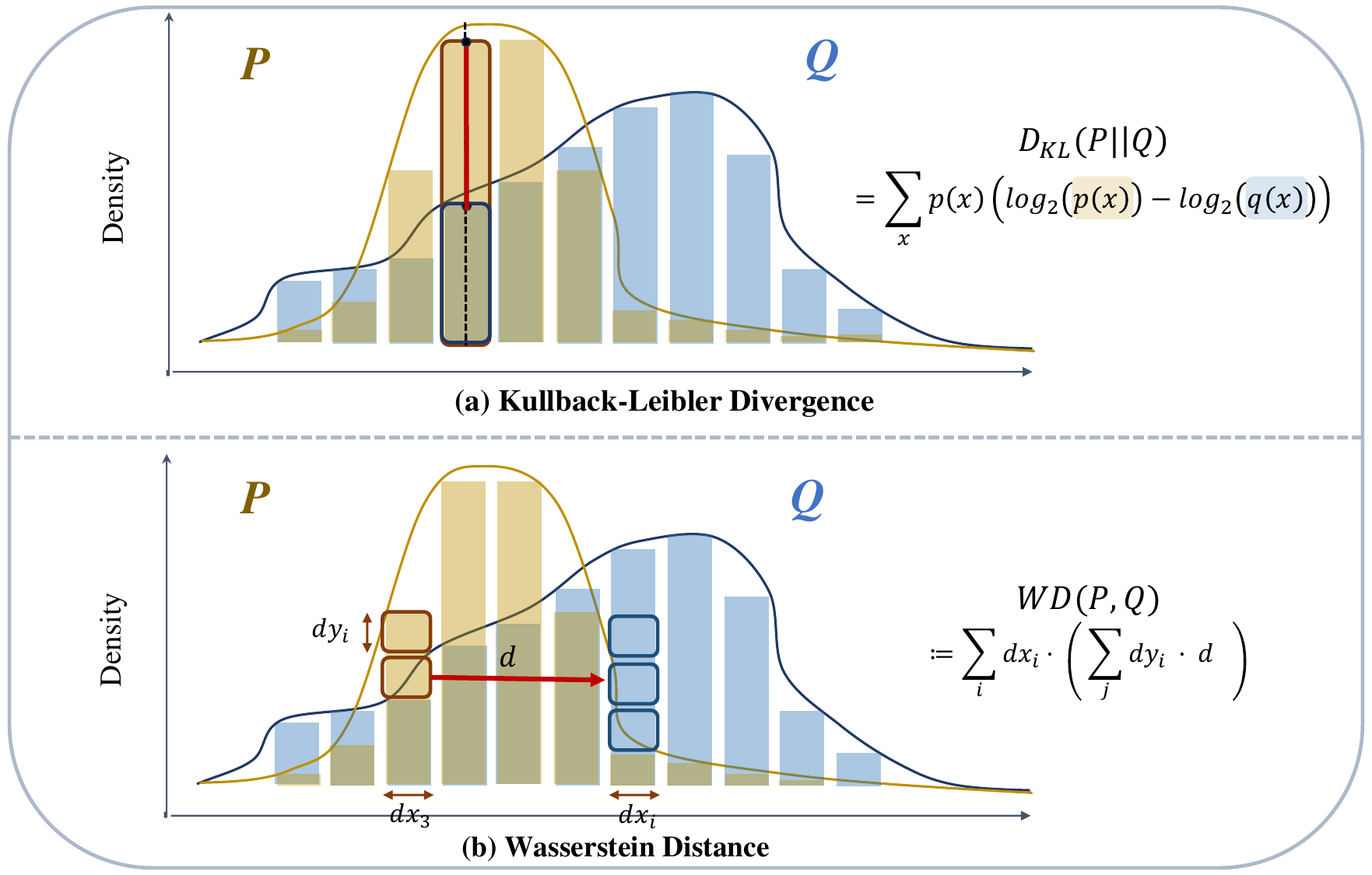}
\vspace{-1.2em}
\caption{Distribution alignment comparison diagram of Kullback-Leibler (KL) divergence and Wasserstein distance. $P$ and $Q$ denote two distinct distributions, representing the distribution of two rPPG’s power spectral densities after undergoing softmax transformation. (a) illustrates the vertical probability dependence of KL divergence, where p(x) and q(x) represent the densities in the distributions $P$ and $Q$. And (b) demonstrates the global advantages of the lateral optimal transmission of Wasserstein distance, where $d$ represents the distance to be moved in the adaptation process. }
\label{fig:WD loss}
\vspace{-1.2em}
\end{figure}

In~\cite{yu2022physformer}, Kullback-Leibler (KL) divergence~\cite{gao2017deep} is used as the label distribution loss to align the output signal with the Gaussian distribution of heart rate label. Nonetheless, as depicted in Fig. \ref{fig:WD loss} (a), KL divergence primarily emphasizes point-to-point probability deviation, overlooking the structural information encompassing the entire distribution. Conversely, the Wasserstein distance (WD)~\cite{ramdas2017wasserstein} offers a more comprehensive measure, addressing not only the deviation between individual probabilities, but also capturing differences across the overall distributions, as depicted in Fig. \ref{fig:WD loss} (b). Thus, this method can more accurately delineate the actual disparity among the frequency general density distributions associated with rPPG signals.

\begin{table*}[t]
\small
		\centering
	\caption{Cross-dataset testing between UBFC and PURE. The result shown with a  \fcolorbox{white}{gray!25}{light grey} background represent a cross-domain from UBFC to PURE. And the result shown with a 
\fcolorbox{white}{gray!50}{dark grey} background represent a cross-domain from PURE to UBFC. Best results are marked in bold.}
 \vspace{-0.3em}
\label{tabel:UBFC_PURE}
	\renewcommand\arraystretch{0.9}
 \setlength{\tabcolsep}{3mm}
	 {\begin{tabular}{@{}cccc|cccccc@{}}
		\toprule
		\multirow{4}{*}{\textbf{Types}} & \multirow{4}{*}{\textbf{Method}} & \multirow{4}{*}{\textbf{Source data}} & \multirow{4}{*}{\textbf{Train data}} & \multicolumn{6}{c}{\multirow{1}{*}{\textbf{Test data}}}\\
  \cmidrule(lr){5-10} & & & & \multicolumn{3}{c}{\multirow{1}{*}{\textbf{UBFC}}} & \multicolumn{3}{c}{\multirow{1}{*}{\textbf{PURE}}}
  \\   \cmidrule(lr){5-7} \cmidrule(lr){8-10}
     & & & 
	 	&  MAE & RMSE & R    & MAE & RMSE & R   \\ 
   \cmidrule(r){1-10} 
   \
    \multirow{4}{*}{\textbf{\rotatebox{90}{Traditional}}}
   & \multirow{2}{*}{CHROM~\cite{verkruysse2008remote}} 
   & UBFC & - & 2.36 & 9.23 & 0.87 & \cellcolor{gray!25}- & \cellcolor{gray!25}- & \cellcolor{gray!25}- \\
    & & PURE & - & \cellcolor{gray!50}3.44 & \cellcolor{gray!50}4.61 & \cellcolor{gray!50}0.97 & 0.75 & 2.23 & 0.99 \\
    \cmidrule(r){3-10} 
     & \multirow{2}{*}{POS~\cite{wang2016algorithmic}} 
    & UBFC & - & 2.11 & 9.11 & 0.87 & \cellcolor{gray!25}- & \cellcolor{gray!25}- & \cellcolor{gray!25}- \\
    & & PURE & - & \cellcolor{gray!50}2.44 & \cellcolor{gray!50}6.61 & \cellcolor{gray!50}0.94 & 0.80 & 4.11 & 0.98 \\
\cmidrule(r){1-10} 
    \
    \multirow{10}{*}{\textbf{\rotatebox{90}{Supervised}}}
   & \multirow{2}{*}{PhysNet~\cite{yu2019remote1}} 
   & UBFC & - & 0.75 & 1.47 & 0.99 & \cellcolor{gray!25}3.81 & \cellcolor{gray!25}- & \cellcolor{gray!25}0.87 \\
    & & PURE & - & \cellcolor{gray!50}7.02 & \cellcolor{gray!50}- & \cellcolor{gray!50}0.60 & 0.99 & 5.22 & 0.93 \\
    \cmidrule(r){3-10} 
     & \multirow{2}{*}{Dual-GAN~\cite{lu2021dual}} 
    & UBFC & - & 0.44 & 0.67 & 0.99 & \cellcolor{gray!25}- & \cellcolor{gray!25}- & \cellcolor{gray!25}- \\
    & & PURE & - & \cellcolor{gray!50}0.74 & \cellcolor{gray!50}1.02 & \cellcolor{gray!50}0.99 & 0.82 & 1.31 & 0.99 \\
    \cmidrule(r){3-10} 
     & \multirow{2}{*}{PulseGAN~\cite{yin2022pulsenet}} 
    & UBFC & - & 1.19 & 2.10 & 0.98 & \cellcolor{gray!25}- & \cellcolor{gray!25}- & \cellcolor{gray!25}- \\
    & & PURE & -  & \cellcolor{gray!50}2.09 & \cellcolor{gray!50}4.42 & \cellcolor{gray!50}0.97 & - & - & - \\
    \cmidrule(r){3-10} 
     & \multirow{2}{*}{Contrast-Phys+ (100\%)~\cite{sun2024contrast}} 
    & UBFC & - & 0.50 & 1.15 & 0.99 & \cellcolor{gray!25}0.89 & \cellcolor{gray!25}2.26 & \cellcolor{gray!25}0.98 \\
    & & PURE & -  & \cellcolor{gray!50}0.61 & \cellcolor{gray!50}2.02 & \cellcolor{gray!50}1.98 & 0.44 & 1.49 & 0.99 \\
\cmidrule(r){1-10} 
   \
    \multirow{8}{*}{\textbf{\rotatebox{90}{Source-free DA}}}
   & \multirow{2}{*}{SHOT~\cite{liang2020we}} 
   & UBFC & PURE & 0.58 & 1.22 & 0.99 & \cellcolor{gray!25}1.17 & \cellcolor{gray!25}2.89 & \cellcolor{gray!25}0.97 \\
    & & PURE & UBFC & \cellcolor{gray!50}0.58 & \cellcolor{gray!50}1.22 & \cellcolor{gray!50}0.99 & 0.83 & 2.60 & 0.98 \\
    \cmidrule(r){3-10} 
     & \multirow{2}{*}{CoWA~\cite{lee2022confidence}} 
    & UBFC & PURE & 0.75 & 1.47 & 0.99 & \cellcolor{gray!25}1.22 & \cellcolor{gray!25}2.98 & \cellcolor{gray!25}0.97 \\
    & & PURE & UBFC & \cellcolor{gray!50}0.83 & \cellcolor{gray!50}1.53 & \cellcolor{gray!50}0.99 & 0.50 & 1.73 & 0.99 \\
    \cmidrule(r){3-10} 
     & \multirow{2}{*}{\textbf{SFDA-rPPG (Ours)}} 
    & UBFC & PURE & 10.92 & 22.30 & 0.29 & \cellcolor{gray!25}\textbf{0.33} & \cellcolor{gray!25}\textbf{1.15} & \cellcolor{gray!25}\textbf{0.99} \\
    & & PURE & UBFC& \cellcolor{gray!50}\textbf{0.41} & \cellcolor{gray!50}\textbf{1.08} & \cellcolor{gray!50}\textbf{0.99} & 0.56 & 1.76 & 0.99 \\
  \bottomrule
	\end{tabular}}
 \vspace{-0.9em}
\end{table*}

Easy to understand, we can assume that there are two piles of stones, and the minimum ``effort" required to stack the first pile of stones into the second pile is the so-called Wasserstein distance. More specifically, this ``effort" or ``workload" can be quantified as the sum of the products of the distance and quantity that the stone must move. For two continuous probability distributions $P$ and $Q$, WD can be defined as,

\begin{equation} \small
W D(P, Q)=\inf _{\gamma \in \Pi(P, Q)} \int_{X \times Y}|x-y| d \gamma(x, y) ,
\end{equation}
where $\Pi(P, Q)$ represents the set of all possible transmission plans to convert $P$ to $Q$, $\gamma$ is one of the specific transmission plans, $|x-y|$ represents the cost of moving from point $x$ to point $y$ in space, and the $\inf$ represents the minimum value taken out of all possible transmission plans.

For rPPG signals, time-domain WD~\cite{sun2022non} calculation can be performed directly to leverage the timing of systolic peaks from contact PPG as labels for model training. However, rPPG signals in time-domain contain too much noise interference, while frequency domain analysis can more easily remove some noise interference by filtering. So we can get the power spectral density of rPPG signals to calculate a more robust alignment loss, and we refer to it as Frequency-domain Wasserstein Distance (FWD) loss. We can express the PSDs of two rPPG signals as P and Q. For these point set distributions composed of real numbers, the transmission plan can also be expressed as:
\begin{equation} \small
FWD(P,Q)=\min_{T\in\Pi} \sum_{i,j} T_{ij} \cdot d(p_i,q_j) ,
\end{equation}
where $T$ is a transportation plan matrix, where $T_{i,j}$ represents the probability quality of transportation from point $p_i$ to $q_j$. $d(p_i,q_j)$ is the distance between $p_i$ and $q_j$. In the case of one-dimensional signal, we simplify the distance to absolute distance. $\Pi$ is the set of all possible transportation plans that satisfy the marginal distribution constraints from $P$ to $Q$.

Specifically, we use the cumulative distribution function CDF~\cite{park2018fundamentals} to find the effort needed for this one-dimensional most suitable transportation plan. Although there may be numerous optimal plans, there exists only one optimal result. Regardless of its specific plan, we only need to calculate the final result, so the transmission FWD can be expressed as:
\begin{equation} \small
FWD(P,Q)=\sum_{i} |CDF_i(P)- CDF_i(Q)|,
\end{equation}
where $CDF_i$ represents the $i$-th value after distribution accumulation.


\subsection{Overall Training and Prediction Procedure}
\label{sec:DA}

\label{sec:wdloss}

\begin{table}[t]
		\centering
	\caption{One-shot evaluations between UBFC and PURE.}
  \vspace{-0.3em}
  \label{tabel:1-shot}
	\renewcommand\arraystretch{1}
 \setlength{\tabcolsep}{2.5mm}
	\resizebox{0.48\textwidth}{!} {\begin{tabular}{@{}p{1.6cm}cccccccc@{}}
		\toprule
		\multirow{2}{*}{\textbf{Method}}  & \multicolumn{2}{c}{\textbf{Data}} & \multicolumn{3}{c}{\textbf{UBFC}} & \multicolumn{3}{c}{\textbf{PURE}} 
        \\ \cmidrule(lr){2-3}\cmidrule(lr){4-6}\cmidrule(lr){7-9}  
		& Source & Target & MAE & RMSE & R    & MAE & RMSE & R   \\ \cmidrule(r){1-9} 
        \multirow{2}{*}{PhysNet~\cite{yu2019remote1}} & UBFC & - & 0.75  & 1.47  & 0.99  & \cellcolor{gray!25}3.81  & \cellcolor{gray!25}- & \cellcolor{gray!25}0.87   \\
             & PURE & - & \cellcolor{gray!50}7.02   & \cellcolor{gray!50}-  & \cellcolor{gray!50}0.60  & 0.99  & 5.22 & 0.93        \\
  \cmidrule(r){1-9}
            \multirow{2}*{SHOT~\cite{liang2020we}}
                & UBFC & PURE & 3.83   & 11.06  & 0.57  & \cellcolor{gray!25}0.33  & \cellcolor{gray!25}1.15 & \cellcolor{gray!25}0.99        \\
                 & PURE & UBFC & \cellcolor{gray!50}1.08   & \cellcolor{gray!50}1.87  & \cellcolor{gray!50}0.98  & 0.56  & 2.00 & 0.98        \\
  \cmidrule(r){1-9} 
        \multirow{2}*{CoWA~\cite{lee2022confidence}}
             & UBFC & PURE  & 0.75   & 1.47  & 0.99  & \cellcolor{gray!25}1.33  & \cellcolor{gray!25}3.06 & \cellcolor{gray!25}0.97        \\
            & PURE & UBFC & \cellcolor{gray!50}0.92   & \cellcolor{gray!50}1.78  & \cellcolor{gray!50}0.98  & 0.50  & 1.73 & 0.99        \\
  \cmidrule(r){1-9} 
  
 \textbf{SFDA-rPPG}
             & UBFC & PURE   & 0.67  & 1.41  & 0.99  & \cellcolor{gray!25}\textbf{0.28} & \cellcolor{gray!25}\textbf{1.11} & \cellcolor{gray!25}\textbf{0.99}\\
           \textbf{(Ours)}& PURE & UBFC & \cellcolor{gray!50}\textbf{0.58}   & \cellcolor{gray!50}\textbf{1.22}  & \cellcolor{gray!50}\textbf{0.99}  & 0.44  & 1.70 & 0.99  \\
  \bottomrule
	\end{tabular}}
 \vspace{-1.3em}
\end{table}

In the pre-training stage, we adopt a supervised learning method. After the source domain data passes through PhysNet~\cite{yu2019remote1}, the predicted rPPG signal $f_{source}$ as well as its power spectral density $F_{source}$ can be obtained. Finally, we use FWD loss to align it with the power spectral density $F_{label}$ of the label,
\begin{equation} 
\mathcal{L}_{\text{source}} = \text{FWD}(\text{softmax}(F_{\text{source}}), \text{softmax}(F_{\text{label}})).
\end{equation}

We use $\mathcal{L}_{source}$ to get the pre-training model weight $H_w$ in source domain, and initialize the encoders $E$, $E_s$, $E_t$ and decoder $\mathcal{D}$. 

In the adaptive stage, we adopt a three-branch structure. The middle branch directly uses the original data to get the rPPG signal $f = \mathcal{D} (E (X _ i^T))$ through the model, and then gets its power spectral density $F$. We use FWD loss to adapt our model,
\begin{equation} \small
\label{eq:consistencyloss}
\begin{aligned}
\mathcal{L}_{\text{adapting}} = & \text{FWD}(\text{softmax}(F), \text{softmax}(F_s)) \\
    & + \text{FWD}(\text{softmax}(F), \text{softmax}(F_t)).
\end{aligned}
\end{equation}

During inference, we do not use all three branches simultaneously. We only use the middle branch, as it is the core branch containing the original dataset, which includes the knowledge obtained through consistency learning with the two augmented branches. So we get the predicted rPPG signal through $ f_s = \mathcal{D}(E_s(X_s^T)) $. 

\section{Experiment}
\label{sec:experiment}

\subsection{Datasets and Metrics}
\textbf{Datasets.}\quad The proposed SFDA-rPPG method has been evaluated on three public datasets of PURE \cite{stricker2014non}, UBFC-rPPG \cite{bobbia2019unsupervised}, and COHFACE \cite{heusch2017reproducible}. We will test the intra-domain and cross-domain performance between UBFC and PURE. In addition, we will test the performance from UBFC and PURE to COHFACE because COHFACE is a more realistic and complex scenario than UBFC and PURE.

\begin{table}[t]
		\centering
	\caption{ Ablation results of different branching settings.}
  \vspace{-0.3em}
 \label{tabel:branch}
	\renewcommand\arraystretch{1}
 \setlength{\tabcolsep}{1mm}
	\resizebox{0.48\textwidth}{!} {\begin{tabular}{@{}p{1.6cm}ccccccccc@{}}
		\toprule
		\multirow{2}{*}{\textbf{Method}} & \multirow{2}{*}{\textbf{Setting}} & \multicolumn{2}{c}{\textbf{Data}} & \multicolumn{3}{c}{\textbf{UBFC}} & \multicolumn{3}{c}{\textbf{PURE}} 
        \\ \cmidrule(lr){3-4}\cmidrule(lr){5-7}\cmidrule(lr){8-10}  
	&	& Source & Target & MAE & RMSE & R    & MAE & RMSE & R   \\ \cmidrule(r){1-10} 
 \multirow{2}{*}{PhysNet}&  \multirow{2}*{0-shot}    & UBFC & - & 0.58   & 1.22  & 0.99  &\cellcolor{gray!25} 1.06  &\cellcolor{gray!25} 2.93 &\cellcolor{gray!25} 0.97        \\
 & & PURE & - & \cellcolor{gray!50}0.83   & \cellcolor{gray!50}1.73  &\cellcolor{gray!50} 0.99  & 0.22  & 1.10 & 0.99        \\
 \cmidrule(r){1-10}
 \multirow{4}*{Only Temporal}
            &  \multirow{2}*{1-shot}    & UBFC & PURE & 3.83   & 11.06  & 0.57  & \cellcolor{gray!25}0.33  & \cellcolor{gray!25}1.15 & \cellcolor{gray!25}0.99        \\
           & & PURE & UBFC & \cellcolor{gray!50}1.08   &\cellcolor{gray!50} 1.87  & \cellcolor{gray!50}0.98  & 0.56  & 2.00 & 0.98        \\
		& \multirow{2}*{ALL}    & UBFC & PURE & 5.83   & 15.38	&0.17	&\cellcolor{gray!25}0.28	&\cellcolor{gray!25}1.11	&\cellcolor{gray!25}0.99       \\
       & & PURE & UBFC &\cellcolor{gray!50} 0.67   &\cellcolor{gray!50} 1.41  &\cellcolor{gray!50} 0.99  & 0.78  & 2.40 & 0.98        \\
       
  \cmidrule(r){1-10} 
  
 \multirow{4}*{Only Spatial}
            &  \multirow{2}*{1-shot}  & UBFC & PURE  & 1.50   & 3.11  & 0.96  & \cellcolor{gray!25}0.50  &\cellcolor{gray!25} 1.97 &\cellcolor{gray!25} 0.98        \\
           & & PURE & UBFC &\cellcolor{gray!50} 0.83   &\cellcolor{gray!50} 2.00  & \cellcolor{gray!50}0.99  & 2.39  & 9.69 & 0.70        \\
		& \multirow{2}*{ALL} & UBFC & PURE   & 5.83   & 15.38	&0.17	&\cellcolor{gray!25}0.33	&\cellcolor{gray!25}1.15	&\cellcolor{gray!25}0.99       \\
       && PURE & UBFC  &\cellcolor{gray!50} 0.58   & \cellcolor{gray!50}1.22  & \cellcolor{gray!50}0.99  & 0.83  & 2.60 & 0.98        \\

       \cmidrule(r){1-10} 
  
 \multirow{3}*{\textbf{SFDA-rPPG}}
            &  \multirow{2}*{1-shot} & UBFC & PURE   & 0.67   & 1.41  & 0.99  & \cellcolor{gray!25}\textbf{0.28} & \cellcolor{gray!25}\textbf{1.11} & \cellcolor{gray!25}\textbf{0.99}        \\
           & & PURE & UBFC & \cellcolor{gray!50}0.58   & \cellcolor{gray!50}1.22  & \cellcolor{gray!50}0.99  & 0.44  & 1.70 & 0.99        \\
           \multirow{1}*{\textbf{(Ours)}}
		& \multirow{2}*{ALL}   & UBFC & PURE & 10.92   & 22.30	&0.29	&\cellcolor{gray!25}0.33	&\cellcolor{gray!25}1.15	&\cellcolor{gray!25}0.99       \\
       &  & PURE & UBFC& \cellcolor{gray!50}\textbf{0.42}   & \cellcolor{gray!50}\textbf{1.08}  &\cellcolor{gray!50} \textbf{0.99}  & 0.56  & 1.76 & 0.99        \\
  \bottomrule
	\end{tabular}}
  \vspace{-1.8em}
\end{table}

PURE~\cite{stricker2014non} dataset consists of 10 subjects and each subject was recorded a 1-minute video under 6 scenarios. The videos were captured at a frame rate of 30 Hz with a cropped resolution of 640x480 pixels.

UBFC-rPPG~\cite{bobbia2019unsupervised} dataset has 42 videos from 42 subjects. The video was captured at 30fps with a resolution of 640x480 in uncompressed RGB format.

COHFACE~\cite{heusch2017reproducible} has 160 one-minute videos of 40 subjects. The heart rate and breathing rate of the recorded subjects are synchronized with the videos. The videos have been recorded at a resolution of 640x480 pixels and a frame rate of 20Hz.

\textbf{Metrics.}\quad Following the methodology outlined in~\cite{sun2022contrast, yu2019remote1} in the field, this study employs mean absolute error (MAE), root mean square error (RMSE), and Pearson correlation coefficient (R) as evaluation metrics. Notably, for MAE and RMSE, lower values indicate reduced error margins, whereas for R, values approaching 1.0 signify diminished error. Among them, both MAE and RMSE are measured in terms of bpm (beats per minute). And for convenience, these units will not be repeated in the subsequent tables or analyses.

\subsection{Experimental Setup}
\label{sec:imp}

In this investigation, the MTCNN algorithm~\cite{zhang2016joint} is employed for detecting facial regions and subsequent cropping within video frames. During the pre-training process, face clips represented as 3×300×128×128 (channels×time×height×width) matrices are utilized to pre-train the model and acquire initial weights. Following this, the pre-trained architecture is disassembled to initialize encoders and decoders specific to the target domain. For the primary training phase, inputs comprised 3×210×128×128 (channels×time×height×width) face clip matrices. Training is conducted utilizing the AdamW optimizer~\cite{liu2023efficientphys} across both pre-training and training stages, across 30 epochs on four NVIDIA TITAN X GPUs, utilizing a learning rate of e-5. In the evaluation phase, following the protocol described in~\cite{sun2022contrast}, each video is segmented into 30-second intervals with heart rate measurements obtained for each segment.

\begin{table}[t]
		\centering
	\caption{Cross-dataset testing from UBFC or PURE to COHFACE.}
  \vspace{-0.3em}
 \label{tabel:COHFACE}
	\renewcommand\arraystretch{1.3}
 \setlength{\tabcolsep}{1mm}
	\resizebox{0.50\textwidth}{!} {\begin{tabular}{@{}cccccccc@{}}
		\toprule
		\multirow{3}{*}{\textbf{Types}} & \multirow{3}{*}{\textbf{Method}}   & \multicolumn{3}{c}{{\textbf{UBFC} $\rightarrow$ \textbf{COHFACE}}} & \multicolumn{3}{c}{{\textbf{PURE} $\rightarrow$ \textbf{COHFACE}}} 
        \\ \cmidrule(lr){3-5}\cmidrule(lr){6-8}  
		& & MAE & RMSE & R    & MAE & RMSE & R   \\ \cmidrule(r){1-8} 
      \multirow{3}{*}{\textbf{Supervised}} 
        & EfficientPhys~\cite{liu2023efficientphys} & 27.22  & 29.27  & 0.01  & 27.34  & 29.57 & -   \\
        & PhysFormer~\cite{yu2022physformer} & 21.27   & 26.84  & 0.03  & 17.74  & 20.52 & -        \\
        & PhysNet~\cite{yu2019remote1} & 12.62   & 19.94  & \textbf{0.23}  & 13.96  & 20.36 & 0.002        \\
  \cmidrule(r){1-8}
      \multirow{3}{*}{\textbf{SFDA}} 
        & SHOT~\cite{liang2020we} & 16.78  & 20.19  & 0.08  & 18.91  & 21.96 & 0.01   \\
        & CoWA~\cite{lee2022confidence} & 23.91   & 27.63  & -  & 15.03  & 18.43 & 0.08        \\
        & \textbf{SFDA-rPPG (Ours)} & \textbf{12.38}   & \textbf{15.28}  & 0.04  & \textbf{12.69}  & \textbf{15.80} & \textbf{0.15}        \\
            
  \bottomrule
	\end{tabular}}
\end{table}

\subsection{Cross-Dataset Testing between UBFC and PURE}

As shown in Table~\ref{tabel:UBFC_PURE}, we conduct cross-domain experiments using the UBFC and PURE datasets to assess the cross-domain performance of our method. Specifically, before conducting evaluations on both the two datasets, we separately pre-train our model on one of them with label and then adapt it on another dataset without label. Additionally, we train the SHOT~\cite{liang2020we} and CoWA~\cite{lee2022confidence} models of SFDA for cross-domain performance assessment. We also show the cross-domain performance after training on the source domain on traditional supervised approaches~\cite{lee2022confidence}. From the results, our method is the best in cross-domain. Under the training of unlabeled target domain, the MAE of PURE data set can reach 0.33, which is very close to the MAE of supervised PURE learning under PhysNet. The results demonstrate that our approach outperforms most of the methods, displaying superior cross-domain performance than the traditional SFDA clustering method.

\begin{table}[t]
\small
	\centering
	\caption{Comparison of loss ablation on PhysNet~\cite{yu2019remote1}. We use Negative Pearson loss~\cite{yu2019remote} by default, combined with the loss marked \textbf{\CheckmarkBold} under Method.}
  \vspace{-0.3em}
 \label{tabel:loss_ablation}
	\renewcommand\arraystretch{1}
	\setlength{\tabcolsep}{0.35mm}
	{\begin{tabular}{@{}ccc|c|cccccc@{}}
			\toprule
			\multicolumn{3}{c|}{\multirow{2}{*}{\textbf{Method}}} & \multirow{4}{*}{\textbf{Train data}} & \multicolumn{6}{c}{\multirow{2}{*}{\textbf{Test data}}}\\  	
			&&&&&&&&&\\[-3pt]
			\cmidrule(lr){1-3}	\cmidrule(lr){5-10}
			 \multirow{2.2}{*}{Frequency} & \multirow{2.2}{*}{KL} &\multirow{2.2}{*}{FWD} & & \multicolumn{3}{c}{\multirow{1}{*}{\textbf{UBFC}}} & \multicolumn{3}{c}{\multirow{1}{*}{\textbf{PURE}}}
			\\
			\cmidrule(lr){5-7} \cmidrule(lr){8-10} 
			& & &
			&  MAE & RMSE & R    & MAE & RMSE & R   \\ 
			\cmidrule(r){1-10}

			\multirow{2}{*}{\textbf{\CheckmarkBold}} & \multirow{2}{*}{\textbf{\XSolidBrush}} & \multirow{2}{*}{\textbf{\XSolidBrush}}
			& UBFC & 0.83 & 1.53 & 0.99 & \cellcolor{gray!25}1.50 & \cellcolor{gray!25}3.87 & \cellcolor{gray!25}0.96 
			\\ & & & PURE & \cellcolor{gray!50}3.33 & \cellcolor{gray!50}12.33 & \cellcolor{gray!50}0.46 & 0.56 & 1.76 & 0.99 
			\\
			\cmidrule(r){1-10} 
			\multirow{2}{*}{\textbf{\XSolidBrush}} & \multirow{2}{*}{\textbf{\CheckmarkBold}} & \multirow{2}{*}{\textbf{\XSolidBrush}}
			& UBFC & 0.75 & 1.47 & 0.99 & \cellcolor{gray!25}4.22 & \cellcolor{gray!25}12.86 & \cellcolor{gray!25}0.54 
			\\ & & & PURE & \cellcolor{gray!50}0.92 & \cellcolor{gray!50}1.78 & \cellcolor{gray!50}0.99 & 0.61 & 2.03 & 0.98 
			\\
			\cmidrule(r){1-10} 
			\multirow{2}{*}{\textbf{\XSolidBrush}} & \multirow{2}{*}{\textbf{\XSolidBrush}} & \multirow{2}{*}{\textbf{\CheckmarkBold}}
			& UBFC & \textbf{0.58} & \textbf{1.22} & \textbf{0.99} & \cellcolor{gray!25}\textbf{1.06} & \cellcolor{gray!25}\textbf{2.93} & \cellcolor{gray!25}\textbf{0.97} 
			\\ & & & PURE & \cellcolor{gray!50}\textbf{0.83} & \cellcolor{gray!50}\textbf{1.73} & \cellcolor{gray!50}\textbf{0.99} & \textbf{0.22} & \textbf{1.05} & \textbf{0.99} 
			
			\\
			

			\bottomrule
	\end{tabular}}
 \vspace{-0.5em}
\end{table}

\subsection{Results from PURE or UBFC to COHFACE}
We utilize the UBFC and PURE datasets and attempt to adapt them to the more complex COHFACE dataset, which comprises facial videos captured under wild lighting conditions. We compare the performance of other SFDA's methods~\cite{liang2020we, lee2022confidence} of cross-domaining and the results of direct cross-domaining in supervised ways. As shown in Table~\ref{tabel:COHFACE}, our findings reveal that although the performance of transferring from both datasets to COHFACE is suboptimal, it yields reduced errors compared to prior methods and achieves some enhancements. However, due to the complex collection environment of COHFACE, it is difficult to learn better knowledge from datasets like UBFC and PURE, which are collected in experimental environments.

\begin{table}[t]
\small
		\centering
	\caption{Ablation of different Backbones. ``Baseline" refers to the version with the bare backbone, while ``SFDA" refers to the version where the backbone is integrated with our proposed SFDA-rPPG framework.}
  \vspace{-0.3em}
 \label{tabel:ablation on backbones}
	\renewcommand\arraystretch{1.5}
 \setlength{\tabcolsep}{1.5mm}
	 {\begin{tabular}{@{}ccccccc@{}}
		\toprule
		 \multirow{4}{*}{\textbf{Method}}  & \multicolumn{6}{c}{\multirow{1}{*}{\textbf{UBFC} $\rightarrow$ \textbf{PURE}}}\\
  \cmidrule(lr){2-7}  & \multicolumn{3}{c}{\multirow{1}{*}{\textbf{Baseline}}} & \multicolumn{3}{c}{\multirow{1}{*}{\textbf{SFDA}}}
  \\   \cmidrule(lr){2-4} \cmidrule(lr){5-7}
     
	 	&  MAE & RMSE & R    & MAE & RMSE & R   \\ 
   \cmidrule(r){1-7} 
   
 \multirow{1}{*}{PhysNet~\cite{yu2019remote1}} 
    & 3.81 & - & 0.87 & \textbf{0.33} & \textbf{1.15} & \textbf{0.99} \\
    \multirow{1}{*}{PhysFormer~\cite{yu2022physformer}} 
      & 5.44 & 14.17 & 0.43 & \textbf{1.33} & \textbf{3.65} & \textbf{0.95} \\
      \multirow{1}{*}{EfficientPhys~\cite{liu2023efficientphys}} 
      & 4.16 & 15.82 & 0.60 & \textbf{2.38} & \textbf{9.98} & \textbf{0.81} \\
  \bottomrule
	\end{tabular}}
 \vspace{-0.9em}
\end{table}

\subsection{One-shot Evaluations between UBFC and PURE}
To ascertain the robust generalization capabilities under in scenarios with limited proprietary data, we evaluate our method for One-shot training. Specifically, we employed all the data from the source domain for pre-training and utilized only single unlabeled video from the target domain for adapting. Subsequently, we conduct testing using UBFC and PURE datasets. As shown in Table~\ref{tabel:1-shot}, we compare the results of SFDA’s methods for One-shot training in the target domain and the outcomes of direct cross-domain method using PhysNet, which is supervised training in the source domain. The analysis demonstrates that our method can achieve commendable cross-domain performance, even when employing only one video from the target domain for training, thus indicating its superior learning and generalization capability.

\subsection{Ablation study}

\subsubsection{Impact of Branchings}

To show the effectiveness of our overall framework, we compare the results produced by our methodology with the detachments of the branch architecture, and also with the direct cross-domain assessment of our pre-trained models. As shown in Table~\ref{tabel:branch}, our approach outperforms other approaches in both 1-shot and ALL settings, establishing its adaptability and effectiveness across diverse conditions. Our findings also indicate that the transition from UBFC to PURE is relatively easy, consequently, the 1-shot learning outcomes surpass the ALL learning outcomes in this scenario. This also explains why the ALL learning outcomes from the sole temporal branch marginally exceed the performance of our approach. Nevertheless, in aggregate, our complete framework is better than the single-branch framework.

\subsubsection{Impact of Loss Functions}
As shown in Table~\ref{tabel:loss_ablation}, we conduct a comparative assessment of various loss functions within PhysNet~\cite{yu2019remote1}, employing the UBFC and PURE datasets for supervised training. To prove the effectiveness of our FWD loss, we compare the losses commonly used in rPPG method, including Frequency loss and KL loss in~\cite{yu2022physformer}. Experiments show that our FWD loss achieves the best performance not only within each dataset but also across datasets. This indicates that our FWD loss method is fully effective and it is expected to be widely used in rPPG field in the future.

\subsubsection{Impact of Backbones}
To verify the generalization of our framework, in addition to PhyNet~\cite{yu2019remote1} as the backbone, we also consider two other backbones, namely Physformer~\cite{yu2022physformer} and EfficientPhys~\cite{liu2023efficientphys}. Specifically, we train the backbones to obtain pre-trained models, and split the backbones of Physformer and EfficientPhys into encoder and decoder parts. Then, we perform source-free domain adaptation based on our proposed three-branch framework. As shown in Table~\ref{tabel:ablation on backbones}, the method incorporating SFDA has significantly improved the cross-domain performance of the original method. This demonstrates that our spatio-temporal consistency framework possesses strong generalization capabilities and can be effectively applied to various backbone architectures.

\begin{figure}
\includegraphics[scale=0.45]{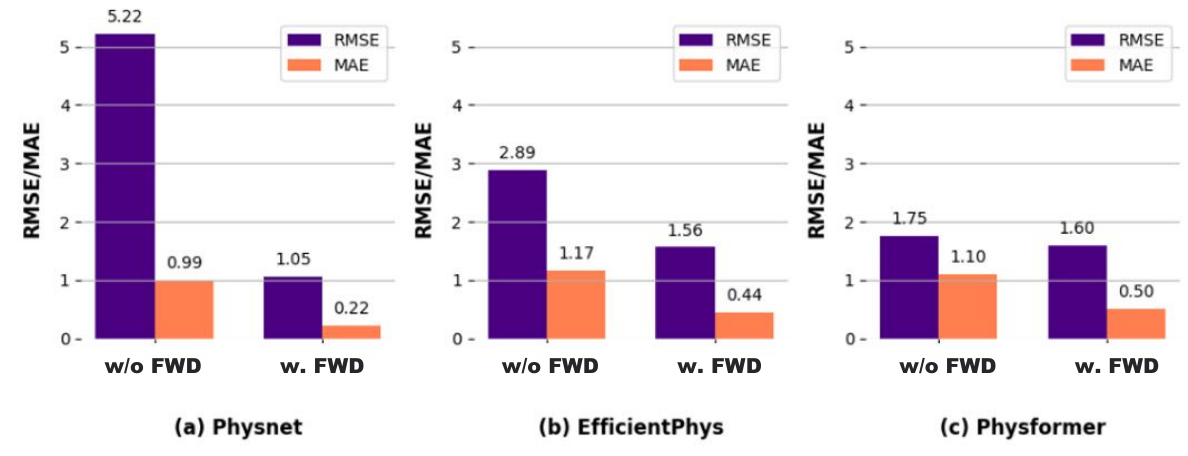}
 \vspace{-1.8em}
\caption{\small{Generalization evaluation of FWD loss on other backbones.  }}
\label{fig:WDonbackbones}
 \vspace{-0.3em}
\end{figure}

\begin{figure}
\centering
\includegraphics[scale=0.85]{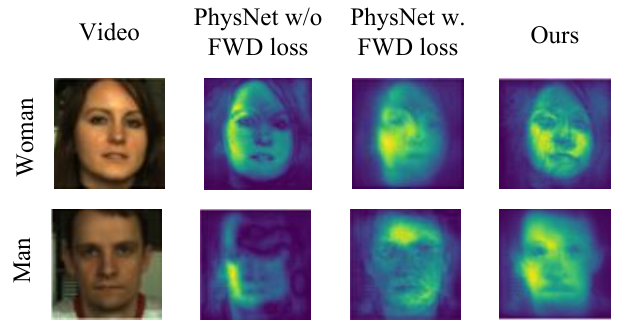}
 \vspace{-1.2em}
\caption{\small{Saliency maps of representative samples on PURE dataset for method with or without FWD loss.}}
\label{fig:Saliencymaps}
 \vspace{-0.8em}
\end{figure}

\subsection{Discussion and Visualization}
\label{sec:Discussion and Visualization}

\subsubsection{Generalization of FWD loss}
\begin{figure}
\centering
\includegraphics[scale=0.5]{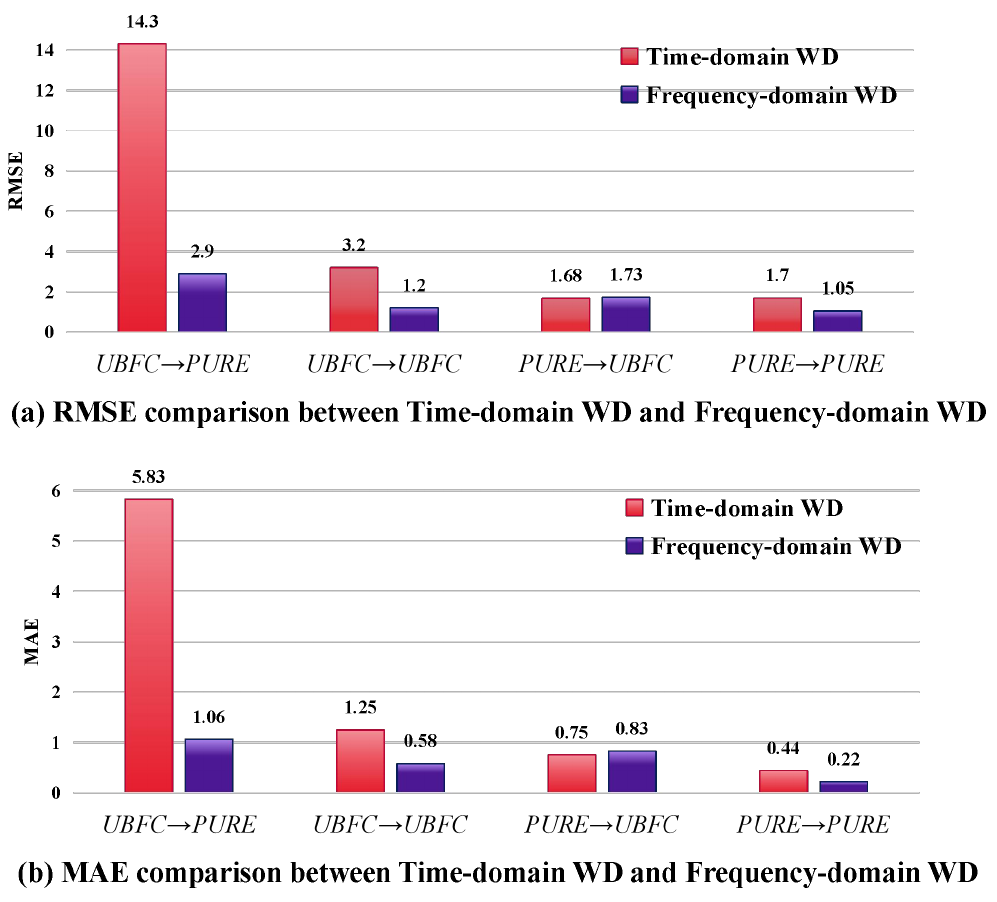}
 \vspace{-0.2em}
\caption{\small{Comparison of time domain and frequency domain results on PhysNet~\cite{yu2019remote1}. Datasets before the ``arrow" refer to the training datasets, while those after the ``arrow" refer to the testing datasets. For example, UBFC→PURE means training on UBFC and testing on PURE.  }}
\label{fig:Time_frequency_WD}
 \vspace{-0.5em}
\end{figure}

We apply the FWD loss to different backbone architectures to ascertain its effectiveness. Besides PhysNet, we extend its application to EfficientPhys and Physformer to assess its robustness across different models. As illustrated in Fig. \ref{fig:WDonbackbones}, experimental results demonstrate that the FWD loss consistently yields favorable generalization across various methods. For example, the MAE on PURE data set is improved from 1.17 to 0.44 under EfficientPhys. In addition, we also found the use of WD in the~\cite{sun2022non}, but we are different. They use WD in the time domain, while we use WD after the signal is converted to frequency domain. We compare the time-domain and frequency-domain results in Fig. \ref{fig:Time_frequency_WD}. We use PhysNet as the basic framework and change loss. The experimental results show that our method is obviously improved. For example, in the case of cross-data sets from UBFC to PURE, our FWD method in frequency domain can reach the result of MAE 1.06, which is better than that in time domain. This indicates that FWD holds promise for adaption by other rPPG models.

\subsubsection{Saliency Maps}
We follow~\cite{sun2022contrast} to generate comparative saliency map of our method and PhysNet~\cite{yu2019remote1} both with and without FWD loss. Specifically, we employ a gradient-based method to calculate the negative Pearson correlation gradient of input video results using fixed model weights, and then get the corresponding saliency map. To evaluate its cross-domain performance from UBFC to PURE, we train PhysNet weights with and without FWD loss on UBFC. Additionally, our SFDA-rPPG is pre-trained on UBFC and then adapts on PURE. We select videos of two subjects from the PURE dataset to conduct a significant graph comparison. As illustrated in Fig. \ref{fig:Saliencymaps}, experimental results demonstrate saliency map of our method activates the majority of skin regions in rPPG data, particularly concentrating on the cheek and forehead with rich rPPG signal. The results of PhysNet with FWD loss are more concentrated than normal PhysNet, again demonstrating the effectiveness of our FWD loss.

\section{CONCLUSIONS}
\label{sec:conclusion}
In this paper, we introduced SFDA-rPPG, a novel framework for source-free domain adaptation in rPPG measurement. Our framework addresses domain shift challenges by incorporating a Three-Branch Spatio-Temporal Consistency Network (TSTC-Net), which effectively reduces the domain gap by learning consistent features across domains. Additionally, we proposed a Frequency-domain Wasserstein Distance (FWD) loss to align the spectrum distributions of rPPG signals, significantly enhancing adaptation performance. SFDA-rPPG presents a breakthrough for cross-domain and privacy-preserving applications in rPPG technology. In future work, we plan to explore pretraining on multiple source domains to further improve source-free domain adaptation in rPPG measurement.





{\small
\bibliographystyle{ieee}
\bibliography{egbib}
}

\end{document}